\PassOptionsToPackage{sort&compress}{natbib}
\documentclass{AR-Style/ar-1col-S2O-edited}
\usepackage[numbers]{natbib}
\usepackage{url}
\usepackage{comment}
\usepackage{subcaption}
\usepackage{graphicx}
\usepackage{amsmath}
\usepackage{siunitx}
\usepackage{hyperref}
\usepackage{float} 
\usepackage{tabularx}
\usepackage{xltabular}
\usepackage{makecell}
\usepackage{multirow}
\usepackage{booktabs}

\usepackage{xspace}

\hypersetup{
    colorlinks=true,
    linkcolor=blue,
    filecolor=blue,      
    urlcolor=blue,
    citecolor=blue,
}

\firstpagenote{To be published in the \emph{Annual Review of Control,
  Robotics, and Autonomous Systems}, Volume 10, 2027}

\begin{document}

% Page header
\markboth{Shilati et al.}{Benchmarking Dexterity of Multifingered Robot Hands}

% Title

\title{\vspace*{-0.7in} \\
Benchmarking Dexterity of Multifingered Robot Hands: A Review and Perspective}
%Authors, affiliations address.

\author{Anthony Shilati$^{1}$, Anunth Ramaswami$^{1}$, Luke Batteas$^{1}$, Sylvia Tan$^{1}$, Anthony Barcio$^{2}$, Sairam Umakanth$^{1}$, Preksha Rao$^{2}$, David McDougall$^{2}$, Landry Graves$^{2}$, Ahmet A. Ozkan$^{3}$, Yunsoo Yoon$^{1}$, Arushi Pradhan$^{2}$, Michael G. Henry$^{4}$, Rohan Kota$^{1}$, Damian Gonzalez$^{1}$, Gray C. Thomas$^{2}$, Gary K. Fedder$^{3}$, J. Edward Colgate$^{1}$, and Kevin M. Lynch$^{1}$
\affil{US National Science Foundation HAND Engineering Research Center\\\
$^1$Center for Robotics and Biosystems, Northwestern Univ., Evanston, IL, USA
\\
$^2$Department of Mechanical Engineering, Texas A\&M Univ., College Station, TX, USA
\\
$^3$Department of Electrical and Computer Engineering, Carnegie Mellon Univ., Pittsburgh, PA, USA
\\
$^4$Department of Mechanical Engineering, Florida A\&M Univ., Tallahassee, FL, USA
\\
Corresponding authors: anthony.shilati@northwestern.edu, kmlynch@northwestern.edu
}
}

%Abstract -- approximately 150 words
\begin{abstract}
Robot hands are a key interface between AI and the physical world, making advances in robotic dexterity essential to realizing the vision of physical AI. While impressive dexterity has been demonstrated with simple grippers, multifingered hands offer the potential for substantially greater versatility, precision, and adaptability in manipulation.

In this review, we survey the state of the art in benchmarking the dexterity of multifingered robot hands. Recognizing dexterity as a complex and multifaceted concept, we present the perspective of the U.S. National Science Foundation HAND Engineering Research Center, with a particular focus on fine in-hand manipulation. We introduce a framework consisting of three benchmark levels that correspond to increasing system complexity, review representative benchmarks at each level, and propose new benchmarks and metrics to address limitations in the literature. More information can be found at~\url{https://hand-erc.github.io/benchmarking/}. 

\end{abstract}

%Keywords, etc.

\maketitle

% Section 1
\section{INTRODUCTION}
\label{sec:Intro}
The US NSF Human AugmentatioN via Dexterity Engineering Research Center (HAND ERC,~\url{https://hand-erc.org}) is a consortium working to revolutionize robotic dexterity. 
Advanced robot dexterity will enable individuals with motor impairments to live fuller, more independent lives; improve the quality of manufacturing jobs while making workers more productive; address changing demographics and labor shortages; make supply chains
less susceptible to disruption; increase the productivity of small and medium enterprises; and democratize access to the benefits of robotics. 

To advance dexterity, we must benchmark it. Benchmarks should measure progress and, ideally, provide actionable feedback to system designers.

But how do we define dexterity?
While there is no universally accepted definition~\cite{ma_dexterity_2011}, our working definition is that dexterity is \emph{the capacity to use contact to effect desired changes to the physical world, or learn properties of the physical world, efficiently and reliably across tasks and conditions}. 
We consider manual dexterity to be a dynamic negotiation between hands, objects being manipulated, and the environment. Hands do not simply dictate motions of manipulanda; instead, they are responsive to, and work in concert with, physical constraints imposed by the task and environment.

Given this broad definition, benchmarking hand dexterity is ill-defined, requiring the imposition of a perspective. In this review paper, we provide the current perspective of the HAND ERC, embracing the fact that this perspective will continue to evolve.

\subsection{The Relevance of In-Hand Manipulation to Dexterity} 

\begin{figure}
    \centering
    \includegraphics[width=\linewidth]{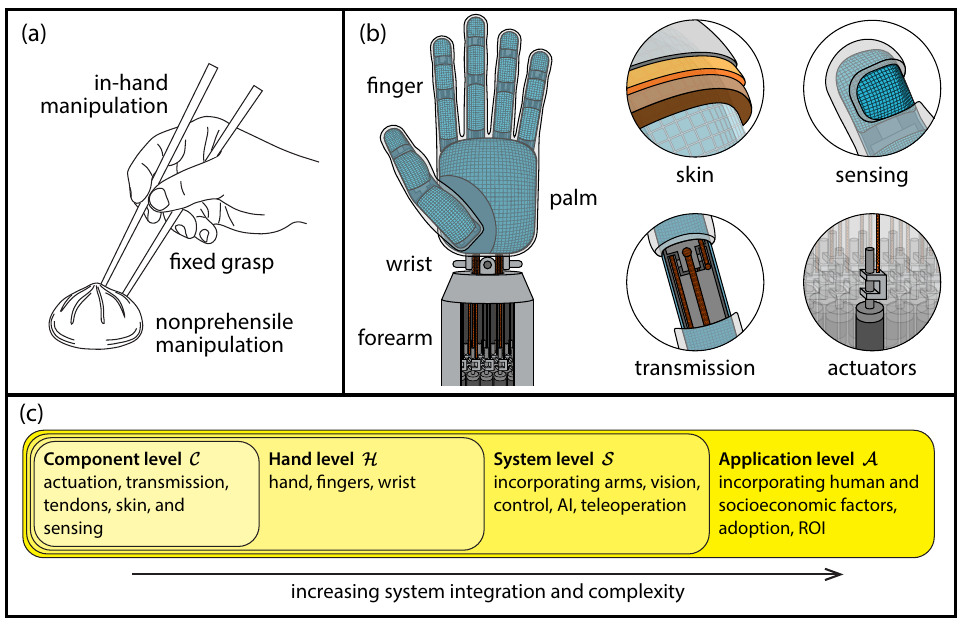}
    \caption{(a) A manipulation task demonstrating direct manipulation of chopsticks and indirect manipulation of the dumpling through tools (the chopsticks). This task also demonstrates the three major types of manipulation: nonprehensile manipulation (pushing the dumpling with chopsticks), fixed grasp manipulation (of the bottom chopstick), and in-hand manipulation (of the top chopstick). (b) Components of a robot hand considered in this review. (c) Four dexterity benchmarking levels, from the lowest component level to the highest application level. The sensitivity of performance to design decisions becomes increasingly complex to estimate for higher-level benchmarks.}
    \label{fig:intro-panels}
\end{figure}

Manipulation may be classified as prehensile (object(s) are grasped) or nonprehensile (no grasp is established, as in pushing) (Figure~\ref{fig:intro-panels}(a)). Within the prehensile category, hands may employ a fixed grasp, e.g., ``pick and place,'' or they may perform in-hand manipulation, where grasped objects move relative to the palm. With a fixed grasp, the primary job of the fingers is to immobilize the object(s) while the arm(s) perform the motion-, force-, or impedance-controlled manipulation. With in-hand manipulation, the fingers --- with their smaller size, lower impedance, and higher control bandwidth --- play a more versatile role in fine control of grasped objects.
Examples of in-hand manipulation include writing with a pen, changing the orientation of a phone in the hand, and many types of tool use.  For hands with fingers with fewer than six degrees of freedom, which includes essentially all robot hands, a simple kinematic analysis shows that in-hand manipulation requires relative motion at contacts, e.g., rolling, sliding, twisting, or some combination.  These relative motions can be achieved by finger motions~\cite{han_dextrous_1998, weng_compliant_2026}, inertial loads~\cite{shi_-hand_2020}, and contact with the environment~\cite{dafle_extrinsic_2014}.

While nonprehensile or fixed-grasp dexterity can be achieved with simple grippers~\cite{mahler_learning_2019, chi_universal_2024}, hooks~\cite{huinink_learning_2016}, and even zero-actuator end-effectors~\cite{lynch_stable_1996}, multifingered hands offer the possibility of far more versatile in-hand manipulation and better exploitation of tools and infrastructure developed for human hands. Accordingly, this paper focuses on benchmarking dexterity with multifingered robot hands, with a particular focus on sensitive control of finger contact forces.

\subsection{Multifingered Hands}
\label{sec:MultifingeredHands}

A multifingered hand (Figure \ref{fig:intro-panels}(b)) has two or more articulated fingers branching from a palm, which itself may be rigid or articulated. Each finger is driven by at least one actuator and may have passive or coupled degrees of freedom. Finger actuators generate torque directly at the joint or are remotized to the palm or forearm, for example, using tendons \cite{salisbury_articulated_1982, jacobsen_design_1986} or linkages \cite{kim_integrated_2021}. 
At-the-joint and remote actuation are called \emph{intrinsic} and \emph{extrinsic} actuation, respectively, in~\cite{intrinsic_extrinsic}.
A wrist, connecting the palm to the forearm, can provide significant rotational mobility to the palm without requiring large, volume-sweeping motions of the arm. Surfaces of a multifingered hand can be covered with tactile sensors, to provide localized feedback, and artificial skin, to provide stabilizing compliance and large contact patches. 

Although we use shorthand terms like forearm, wrist, palm, and finger for multifinger hands, this review is not limited to anthropomorphic robot hands.

\subsection{Dexterity Benchmarking Levels}
In this review, a \emph{benchmark} is a test protocol that yields \emph{scores} for one or more \emph{metrics}, such as success rate, speed of completion, or throughput. Some metrics, such as the number of degrees of freedom of a hand, can be evaluated without performing a test.  

Ultimately, the utility of a dexterous system may be judged by its value in practical applications in manufacturing, home health care, logistics, and other domains. \emph{Application-level} dexterity benchmarks consider socioeconomic factors, human acceptance, return on investment, and other considerations impacting the practical adoption of dexterous robotics. Such considerations are beyond the scope of this paper.  

Our focus begins with the ability of a robot system to perform short-horizon ``atomic'' dexterous manipulation tasks. 
Consider making a peanut butter and jelly sandwich. This is a long-horizon task requiring both dexterity and ``cognition'' (e.g., conditional sequencing or planning), while short-horizon subtasks, like twisting the lid on a peanut butter jar, focus on dexterity. Evaluating a short-horizon manipulation task like jar-capping requires a fully integrated manipulation system, including hands, robot arms, vision, control, AI, possibly teleoperation, and more. Because of the high level of integration needed to perform short-horizon benchmark tasks, we call these \emph{system-level} benchmarks. An example system-level metric is the average time needed to twist a lid on a jar. Most previous work on benchmarking robot dexterity is at the system or long-horizon level.

Benchmarks ideally provide actionable feedback to dexterous system designers. Conceptually, let $\mathcal{D}$ represent a set of scalar-valued design variables and $\mathcal{S}$ represent performance scores on system-level benchmarks. The dependence of $\mathcal{S}$ on $\mathcal{D}$ can be written $\mathcal{S}(\mathcal{D})$.
Then $(\partial \mathcal{S}/\partial \mathcal{D})|_{\mathcal{D}^0}$ could represent system-level performance sensitivity to changes in design relative to an initial design $\mathcal{D}^0$. Given the complexity of the fully integrated system, however, these sensitivities may be difficult to establish. For example, how sensitive is performance in the jar-capping task to the spatial resolution of tactile sensing?

To help address this attribution problem, we define additional lower-level benchmarks, called \emph{hand-level} and \emph{component-level} benchmarks. An example of a component-level metric is the spatial resolution of tactile sensing, as it depends primarily on tactile sensing components of the hand. Examples of hand-level metrics are the bandwidth and accuracy of fingertip force control, which depend on several factors, such as the mechanical design, actuation, and feedback control of the finger. Although a complete dexterous system involves many components in addition to those of the hand (e.g., vision hardware, AI algorithms, etc.), in this review, we focus on hand hardware components.

Component-, hand-, system-, and application-level benchmarks exist on a spectrum of increasing system integration and complexity (Figure~\ref{fig:intro-panels}(c)). We define $\mathcal{C}$ and $\mathcal{H}$ to be the component- and hand-level benchmarking performance. 
To estimate the sensitivity $\partial \mathcal{S}/\partial \mathcal{D}$, it may be easier to estimate analogs of the sensitivities $\partial \mathcal{S} /\partial \mathcal{H}$, $\partial \mathcal{H} /\partial \mathcal{C}$, and $\partial \mathcal{C}/\partial \mathcal{D}$, establishing a sensitivity chain from design decisions to full system performance, e.g., $\partial \mathcal{S}/\partial \mathcal{D} = (\partial \mathcal{S}/\partial \mathcal{H}) (\partial \mathcal{H}/\partial \mathcal{C}) (\partial \mathcal{C}/\partial \mathcal{D})$.\footnote{This type of continuous sensitivity analysis cannot capture the effects of qualitative shifts of strategy, e.g., ``Fosbury flop'' revolutions~\cite{Goldenberg2010}.}

% Section 2
\section{SYSTEM-LEVEL BENCHMARKING}

Existing system-level dexterity benchmarks generally fall into two categories: (1) those derived from human hand-centric grasp and manipulation taxonomies and (2) tasks that are agnostic of the manipulator. In this review, we examine both perspectives and highlight the need for principled hardware-agnostic benchmarks.

\subsection{Taxonomy-Inspired Benchmarks}
Manipulation benchmarks have been developed based on human grasp taxonomies (Section~\ref{sec:handbenchmarks}) and human in-hand manipulation. Elliott and Connolly~\cite{Elliott1984ACO} proposed categorizing in-hand manipulation into 13 movement classes based on finger synergies and sequential movement patterns. Exner \cite{exner1992inhand} developed a taxonomy that decomposes in-hand manipulation into translational and rotational movements, though this framework is more commonly used in clinical contexts. While these taxonomies focus strictly on human in-hand manipulation, Bullock et al.~\cite{Bullock2013AHC} proposed a broader taxonomy that generalizes across tasks and hand morphologies by categorizing hand-object interactions through contact relationships rather than anthropomorphic movement primitives. These frameworks primarily focus on single-hand rigid-body manipulation; coordinated multi-hand and deformable object interaction have been addressed in more recent extensions, such as those proposed by Krebs and Asfour~\cite{krebs2022} and Blanco-Mulero et al.~\cite{BlancoMulero2024}.

These taxonomies have informed several system-level robotic hand benchmarking efforts. Zhou et al.~\cite{Zhou2020} combined the grasp taxonomy of Feix et al.~\cite{Feix2016TheGT} with the in-hand manipulation categories of Bullock et al.~\cite{Bullock2013AHC} to construct 50 benchmark tasks spanning both grasping and in-hand manipulation behaviors. Their evaluation of in-hand manipulation performance relies largely on qualitative descriptions of smoothness and stability. To address this limitation, Coulson et al.~\cite{Coulson2021} proposed a benchmark derived from the manipulation patterns of Elliott and Connolly~\cite{Elliott1984ACO} using the standardized set of YCB objects~\cite{Calli2015} (Figure \ref{fig:system-level}(a)). Their framework combines qualitative success metrics with quantitative measures of average rotation and translation, normalized by finger length to enable fair comparison across robotic hands.

Recently, Liconti et al.~\cite{liconti2026} proposed the POMDAR benchmark, which addresses both grasp-centric and task-centric benchmarking.  POMDAR tests six grasps and twelve atomic manipulation tasks (Figure \ref{fig:system-level}(b)), targeting coverage across the taxonomies of Elliott and Connolly~\cite{Elliott1984ACO}, Feix et al.~\cite{Feix2016TheGT}, and Bullock et al.~\cite{Bullock2013AHC}. Performance is evaluated using a throughput metric that combines task success rate and completion time.

\subsection{Task-Centric Benchmarks}
While these human hand taxonomy-driven benchmarks provide a structured methodology for benchmark design, they inherit anthropomorphic assumptions.
Clinical dexterity assessments provide an alternative framework based on tasks related to activities of daily living. Tests such as Jebsen--Taylor~\cite{jebsen1969objective}, the Southampton Hand Assessment Procedure (SHAP)~\cite{light2002establishing}, the Chedoke Arm and Hand Activity Inventory (CAHAI)~\cite{barreca2004development, barreca2005test}, and ``box and blocks''~\cite{mathiowetz1985adult} evaluate dexterity through standardized functional tasks, emphasizing task completion capability over specific movement patterns. This task-centric perspective provides a foundation for creating embodiment-agnostic dexterous manipulation benchmarks.

\begin{figure}
    \centering
    \includegraphics[width=1.0\linewidth]{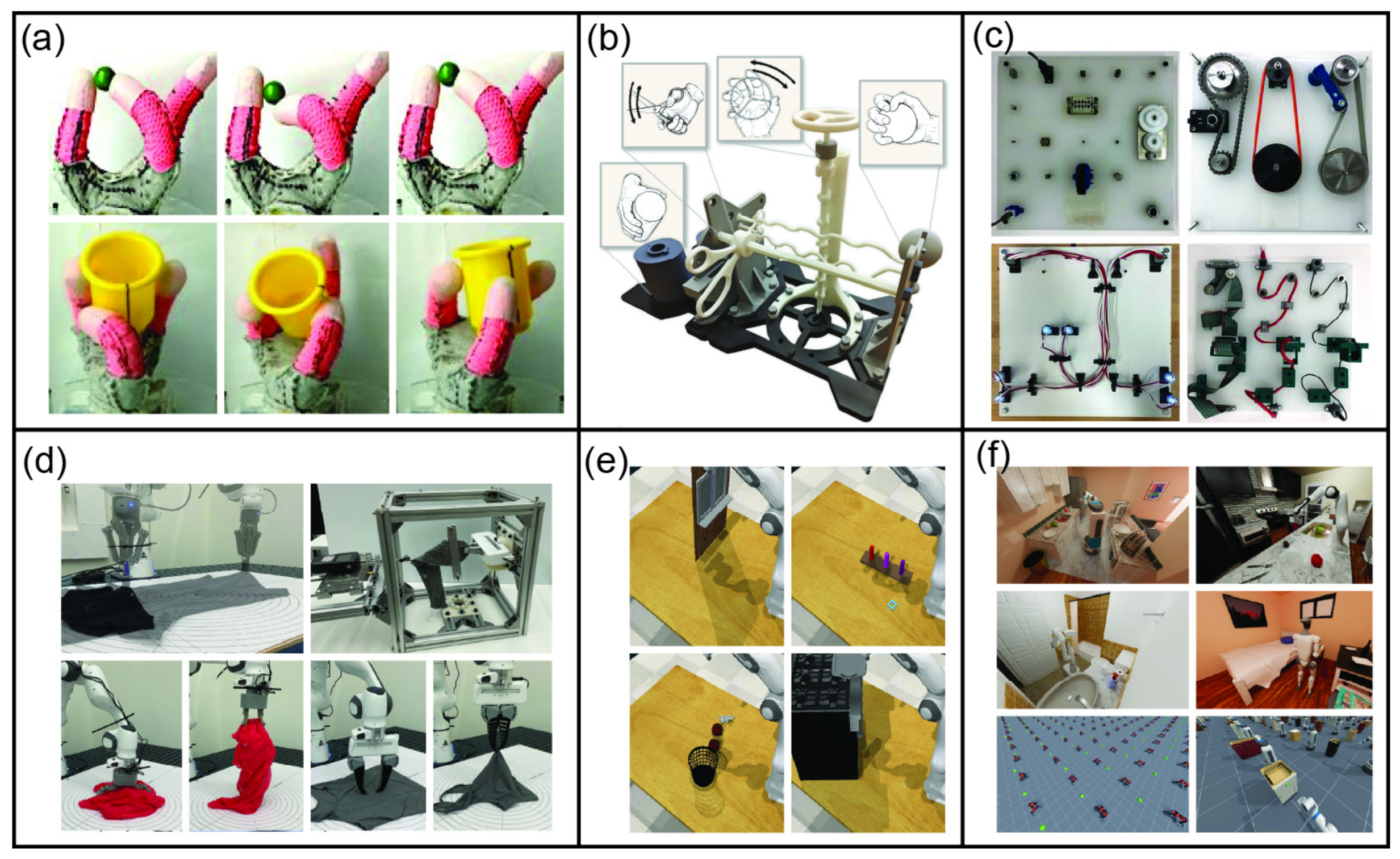}
    \caption{Examples of existing system-level dexterity benchmarks.
    \textbf{Grasp taxonomy-inspired benchmarks}:
    (a) CMU Foam Hand~III performing Elliott and Connolly manipulation patterns (adapted from~\cite{Coulson2021} with permission).
    (b) Equipment for evaluating the POMDAR benchmarks (adapted from~\cite{liconti2026} with permission).
    \textbf{Task-centric domain-specific benchmarks}:
    (c) NIST taskboards (adapted from~\cite{kimble2020} with permission).
    (d) Dyson household clothing benchmark (adapted from~\cite{dysonbench2023} with permission).
    \textbf{Task-centric large-scale simulation benchmarks}:
    (e) RLBench simulations (adapted from \cite{james2019rlbench} with permission).
    (f) ManiSkill3 implementations (adapted from \cite{taomaniskill3} with permission).}
    \label{fig:system-level}
\end{figure}

In this vein, many robot manipulation benchmarks focus on constructing representative task suites for specific manipulation domains. For example, Kimble et al.~\cite{kimble2020} introduced the NIST assembly task boards designed to replicate manufacturing operations such as threading, snap fitting, and gear meshing, evaluating systems using completion time and reliability metrics (Figure \ref{fig:system-level}(c)). Cruciani et al.~\cite{Cruciani2020} proposed an in-hand manipulation benchmark using YCB objects \cite{Calli2015}, with performance measured by error in the final object pose and fingertip locations. Similarly, deformable manipulation benchmarks proposed by Lin et al.~\cite{corl2020softgym}, Chen et al.~\cite{chen2023daxbench}, and Clark et al.~\cite{dysonbench2023} provide task suites tailored to deformable object interaction, employing metrics specific to object deformation outcomes (Figure \ref{fig:system-level}(d)).

Beyond domain-specific benchmarks, the rise of data-driven robot manipulation has motivated the development of large-scale simulation benchmarks designed to evaluate performance across diverse manipulation scenarios. Many of these benchmarks are tied to specific manipulator embodiments, including single-arm parallel-jaw gripper systems \cite{Mu2021ManiSkillGM, james2019rlbench, mclean2025metaworld} (Figure \ref{fig:system-level}(e)), bimanual parallel-jaw gripper systems \cite{Mu_2025_CVPR, chen2025robotwin}, single-arm multifingered hands \cite{Bao2023,dexverse}, and bimanual hand systems \cite{wang2026,Chen2022TowardsHB}. Others, such as ManiSkill3 \cite{taomaniskill3} (Figure \ref{fig:system-level}(f)) and RoboMIND \cite{wu2025robomind}, seek to generalize across many manipulation embodiments and domains. Rather than focusing on a particular niche of robot manipulation, these benchmarks provide broad coverage through diverse manipulanda, manipulation skills (e.g., reaching, pushing, and insertion), and long-horizon interaction scenarios.

A common challenge in constructing such benchmarks is determining which tasks should be included to achieve broad coverage of manipulation capabilities. To address this problem, Wang et al.~\cite{wang2026} propose a data-driven methodology for benchmark construction. Rather than manually defining benchmark tasks, the authors extracted interaction primitives from large-scale robot learning datasets such as Open X-Embodiment \cite{open_x_embodiment_rt_x_2023} and AgiBot World \cite{bu2025agibot_iros} and used a large language model to organize and expand these primitives into a diverse set of manipulation activities. The resulting task suite was then refined through expert review to produce a benchmark suite of 100 representative manipulation tasks, which they call the Great March~100.

While these benchmarks provide broad coverage of manipulation tasks, many are intended to be tested in simulation, which struggles to fully capture sensing noise, contact dynamics, hardware variability, and environmental uncertainty encountered by physical robotic systems. To address these limitations, ManipulationNet~\cite{chen2026manipulationnet} proposes a framework for real-world robot manipulation benchmarking based on reproducible hardware kits (similar to the NIST task boards) and a unified evaluation client. The benchmark currently includes five representative tasks spanning Physical Skills and Embodied Reasoning tracks, evaluating both low-level manipulation capabilities and higher-level embodied intelligence. More importantly, its standardized infrastructure enables new tasks to be incorporated in a consistent and reproducible manner, allowing the benchmark to evolve while maintaining comparability across research groups and robotic platforms. 

The benchmarks above consist primarily of long-horizon tasks requiring both cognition and dexterity.
Recent work toward atomic manipulation benchmarks includes that of 
Rajeswaran et al.~\cite{Rajeswaran2018}, who identified four broad classes of dexterous manipulation, namely object relocation, in-hand manipulation, manipulation of environmental props, and tool use. To represent these categories, they proposed a benchmark suite consisting of four atomic tasks: relocating a sphere, manipulating a pen within the hand, opening a door, and hammering a nail. By selecting a single representative task for each category, the benchmark seeks to capture the core capabilities required for dexterous manipulation while maintaining a compact evaluation suite.

DexBench~\cite{rlwrld2026dexbench} adopts a task-based perspective, characterizing manipulation tasks along six ``orthogonal'' object state complexity (OSC) axes, including geometric complexity, force sensitivity, contact mode complexity, observation incompleteness, deformable complexity, and dynamic/temporal complexity. A particular manipulation task can be assigned a scalar value along each of these six dimensions, forming the ``OSC vector'' for the task. DexBench proposes 56~evaluation cases sorted into 18~task categories, a mix of atomic manipulation tasks and long-horizon tasks, intended to test different regions of the OSC space.

\begin{table}[]
{\raggedright
\arraybackslash
\begin{tabular}{
>{\raggedright\arraybackslash}p{3.5cm}
>{\raggedright\arraybackslash}p{4.2cm}
>{\raggedright\arraybackslash}p{4.1cm}}

\toprule
\textbf{Benchmark category}
& \textbf{Subcategory}
& \textbf{References} \\
\addlinespace[2pt]
\hline
\addlinespace[2pt]

\multirow{2}{2.8cm}{Taxonomy-inspired benchmarks}
& Grasp/movement primitive benchmarks
& \cite{Zhou2020, Coulson2021} \\
\addlinespace[2pt]

& Atomic task benchmarks
& \cite{liconti2026} \\
\addlinespace[3pt]
\hline
\addlinespace[3pt]

\multirow{4}{2.8cm}{Task-centric benchmarks}
& Clinical dexterity tests
& \cite{jebsen1969objective, light2002establishing,
barreca2004development, barreca2005test, mathiowetz1985adult} \\
\addlinespace[2pt]

& Domain-specific benchmarks
& \cite{kimble2020, Cruciani2020, corl2020softgym,
dysonbench2023} \\
\addlinespace[2pt]

& Large-scale simulation benchmarks
& \cite{Mu2021ManiSkillGM, james2019rlbench, mclean2025metaworld,
Mu_2025_CVPR, chen2025robotwin, Bao2023, wang2026,
taomaniskill3, dexverse, wu2025robomind, Chen2022TowardsHB, chen2023daxbench} \\[2pt]

& Atomic task benchmarks
& \cite{Rajeswaran2018, rlwrld2026dexbench, chen2026manipulationnet} \\
\addlinespace[2pt] 
\bottomrule
\end{tabular}
}
\normalsize
\vspace*{0.2in}
\caption{Existing system-level benchmarks.}
\label{tab:sysbench}
\end{table}

\subsection{Perspective}
\label{sec:HighLevelPerspective}

To benchmark the dexterity of our own flagship Dexterity Nexus bimanual research testbed  (DexNex~\cite{DexNexvideo, Buckley2024iros}) as it evolves over the next several years, we have recently defined a set of system-level dexterity benchmark tasks. Our choices were biased toward (1) atomic tasks that test different features of dexterity;
(2) a compact benchmark suite (fewer than 20 tasks); (3) tasks spanning a wide range of difficulty and task characteristics, as discussed below; (4) where possible, tasks with previous support in the literature, as reviewed above (Table~\ref{tab:sysbench}); and (5) tasks that are likely to benefit from in-hand manipulation, beyond fixed-grasp manipulation where the hands' main role is to affix objects to the arms. Although most benchmark tasks are defined independent of the solution hardware or strategy, we expect that in-hand manipulation will lead to better scores for many.

Benchmark tasks were chosen to test different combinations of the following characteristics of dexterous tasks:
\begin{itemize}
    \item direct vs.\ indirect (e.g., tool use) manipulation
    \item rigidity of manipulanda and the environment
    \item size of manipulanda relative to the hands
    \item number of degrees of freedom that must be controlled (e.g., number of objects, deformability)
    \item constraint complexity (number of active equality and inequality environmental constraints, tolerance, rigidity; utility of motion, force, and impedance control) 
    \item constraint change during the task (e.g., establishing and breaking contacts during assembly and disassembly)
    \item quasistatic vs.\ dynamic (e.g., whether the task requires exploitation of dynamics)
\end{itemize}
Each benchmark task consists of a name; a description of the base task, including materials, start state, and success and failure criteria; metrics; a justification for the task based on the descriptors above; and parametric variations of the base task to allow adjusting task difficulty. A summary of the HAND system-level benchmark tasks is given in Table~\ref{table:system-level}. We expect to modify this task set over time as dexterous systems become more capable.
More details can be found at the HAND ERC benchmarking website,~\url{https://hand-erc.github.io/benchmarking/}.

\begin{table}[]
{\raggedright
\arraybackslash
\begin{tabular}{
>{\raggedright\arraybackslash}p{2.8cm}
>{\raggedright\arraybackslash}p{10.5cm}}

\toprule
\textbf{Task name}
& \textbf{Description} \\
\addlinespace[2pt]
\hline
\addlinespace[2pt]

Box and blocks
& Transfer as many small wooden blocks as possible, one at a time,
from a cluttered bin to another bin within 60~seconds
~\cite{mathiowetz1985adult}. \\
\addlinespace[2pt]

Peg-in-hole
& Insert pegs of various shapes and tolerances into the NIST
Assembly Taskboard M1~\cite{kimble2020}. \\
\addlinespace[2pt]

Pick up flat object
& Grasp and lift an M6 washer initially resting flat on a rigid
tabletop~\cite{rlwrld2026dexbench}. \\
\addlinespace[2pt]

Tie a knot
& Tie a string on a knot board into a tight overhand knot. \\
\addlinespace[2pt]

Twist lid on jar
& Pick up a regular mouth Mason jar lid from a table and thread and
tighten it onto a fixed jar. \\
\addlinespace[2pt]

Use screwdriver
& Beginning with a partially threaded screw, grasp a screwdriver and
use it to drive the screw into a threaded hole. \\
\addlinespace[4pt]

Use scissors
& Pick up and use scissors to cut out a circle drawn on a piece of
A4 printer paper. \\
\addlinespace[4pt]

Fasten button
& Insert a button through a buttonhole on a buttoning board and fully
fasten the button. \\
\addlinespace[4pt]

In-hand reorientation
& Change the orientation of a small rectangular prism from one stable grasp to another. \\
\addlinespace[4pt]

Spin a top
& Spin a hand-spun top so that it remains upright as long as possible. \\
\addlinespace[4pt]

Use chopsticks
& Use a pair of chopsticks to grasp small wooden cubes individually
and transfer them into a target container (box and blocks with
chopsticks). \\
\addlinespace[4pt]

Blindly retrieve an object in clutter
& Retrieve a rigid object of a particular shape from among several
distractor objects in a box without exteroceptive sensing
(e.g., vision). \\
\addlinespace[4pt]

Bundle socks
& Bundle two socks by rolling the socks together and tucking one sock
cuff over the other. \\
\addlinespace[4pt]

Zip a zipper
& Zip and unzip a garment zipper on a task board. \\
\addlinespace[4pt]

Paper folding
& Fold a piece of A4 paper in half with a sharp crease. \\
\addlinespace[4pt]

Bundle dowels with a rubber band
& Gather seven dowels into a compact bundle and secure
them by wrapping a rubber band around the bundle twice. \\
\addlinespace[2pt] 
\bottomrule
\end{tabular}
}
\normalsize
\vspace*{0.2in}
\caption{Summary of current HAND system-level benchmarks.}
\label{table:system-level}
\end{table}

% Section 3
\section{HAND-LEVEL BENCHMARKING} 
\label{sec:HandLevel}

With the increasing number of multifingered robot hands coming to market, standard entries are beginning to emerge for robot hand datasheets. Typical entries include weight, number of fingers, number of actuators, total number of degrees of freedom (including passive and coupled degrees of freedom), a characterization of the hand's size and workspace (e.g., range of motion of the joints, Kapandji score measuring the thumb's ability to oppose the fingers~\cite{kapandji_cotation_1986}, and minimum and maximum grip diameter), fingertip position repeatability and accuracy (possibly including joint backlash), speed (such as maximum joint speed, fingertip speed, or open-close cycle frequency), and strength (such as maximum fingertip force, pinch force, grasping force, and/or some measure of payload capacity). These entries largely track established entries for the more mature datasheets of robot arms: weight, degrees of freedom, workspace, repeatability, accuracy, maximum joint speed, and maximum payload.

Unlike robot arms, however, robot hands do their most important work when they are engaged in a dynamic, responsive, and often delicate negotiation with manipulated objects and their environment, e.g., when managing the transition from six degrees of freedom to a single degree of freedom when threading a nut onto a bolt. Such tasks benefit from in-hand manipulation and low-impedance force control. 
In short, along the chain from proximal arm links to the distal links of the fingers directly engaged with the environment, masses and inertias drop significantly, and there is an increasing need for sensitive force control, ``backdrivability,'' and transparency between forces and motions at the fingers and forces and motions measured and controlled at the actuators. 
This motivates the definition of new hand-level metrics to be reported on future hand datasheets.

We begin with a review of hand-level benchmarking and end with our perspective on useful entries for robot hand datasheets of the future.

\label{sec:MediumLevel}
\subsection{Common Benchmarks for the Hand, Wrist, and Fingers}
\label{sec:Hand}
The number of degrees of freedom (DOF) is the minimum number of parameters required to describe a robot's configuration~\cite{lynch_modern_2017}. Salisbury and Craig introduced mobility and connectivity~\cite{salisbury_articulated_1982}, where mobility is the number of DOFs in the hand, and connectivity is the number of relative DOFs between the hand and manipulated object.
The number of DOFs of a robot hand, fingers, and wrist is always reported on datasheets.  

Durability, which includes both impact and fatigue resistance, has received relatively little attention in the literature, no doubt due to the cost of the tests. Negrello et al.\ tested the ability of a robot hand to absorb impact energy using a pendulum apparatus~\cite{negrello_benchmarking_2020}. A force-torque sensor at the base of the hand measures the impact force transmitted through the hand, which is related to its ability to absorb impacts. Similarly, the hand of the DLR hand-arm system was subjected to impact by a 500~g hammer to test robustness~\cite{grebenstein_dlr_2011}, but specific details of the tests were not provided. For commercial robot hands, there are no standard measures of durability, operational lifetime, or serviceability.

The ORCA, RUKA, and LEAP hands were designed as low-cost, robust robotic hands~\cite{christoph_orca_2025, zorin_ruka_2025, shaw_leap_2023} for real-world data collection. For each of these hands, continuous operation tests are used to benchmark repeatability and robustness. Repeatability measures the accumulated positioning error during continuous operation and is assessed by commanding the hand to repeatedly grasp and ungrasp a compliant object~\cite{shaw_leap_2023, christoph_orca_2025}. Grasp endurance is measured by testing how long a hand can statically hold a weight hanging from the fingertips.

\subsection{Hand Benchmarks}
\label{sec:handbenchmarks}
\begin{figure}
    \centering
    \includegraphics[width=4.91in]{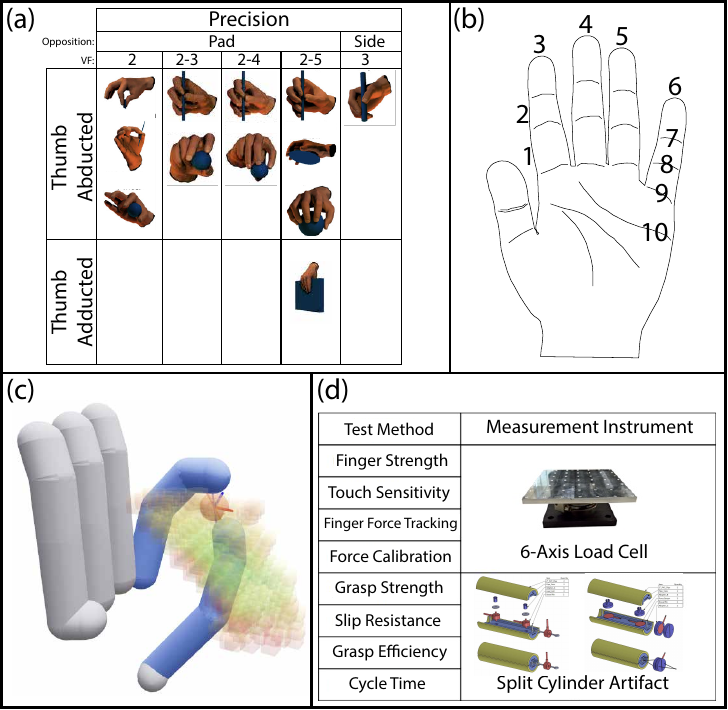}
    \caption{Existing hand-level benchmarks. \textbf{Mobility benchmarks:}
    (a) Twelve of the 33 grasps in the GRASP taxonomy, representing the precision grasp category (adapted from \cite{Feix2016TheGT} with permission). Robot hands are given a score based on how many of the grasps in the GRASP taxonomy they can achieve. 
    (b) A visualization of the 10 points on the hand for the Kapandji score \cite{kapandji_cotation_1986}.  A hand is given a score based on how many of these points the thumb can touch.
    (c) The KaRMA benchmark evaluates a hand's ability to translate or rotate an object in a two-finger pinch grasp for various object locations in the hand's workspace (adapted from \cite{peticco_karma_2026} with permission).
    \textbf{Physical performance benchmarks:}
    (d) Performance measures proposed by NIST and the associated measurement instruments (adapted from \cite{falco_benchmarking_2020} with permission).
    }
    \label{fig:HandLevelReview}
    
\end{figure}

Hand benchmarks can be categorized as grasping and physical performance benchmarks. Most grasping benchmarks compare how many grasps from a human grasp taxonomy a robot hand can achieve~\cite{bridgwater_robonaut_2012}. Cutkosky's grasp taxonomy~\cite{Cutkosky1989OnGC}, rooted in the work of Schlesinger~\cite{schlesinger1919} and Napier~\cite{Napier1956ThePM}, defines sixteen human hand grasp types, primarily categorized as power and precision grasps. Within each of those categories, grasps are further organized by object geometry and hand configuration. Feix et al.~\cite{Feix2016TheGT} expanded on Cutkosky's grasp taxonomy by constructing the GRASP taxonomy (Figure \ref{fig:HandLevelReview}(a)), a set of 33 human hand grasp types.

The anthropomorphic hand assessment protocol (AHAP)~\cite{llop-harillo_anthropomorphic_2019} uses eight of the most common grasps found in activities of daily living and 25 objects from the YCB object set~\cite{Calli2015}. The hand's ability to complete each grasp is scored and a composite grasping ability score is computed. A similar grasping benchmark measures the volumetric grasping capacity of the hand by testing the maximum and minimum diameter cylinders a hand can stably grasp~\cite{kragten_proposal_2010, falco_grasping_2015} . 

Grasp resilience~\cite{negrello_benchmarking_2020, grebenstein_dlr_2011} and grasp pullout strength~\cite{kragten_proposal_2010, falco_grasping_2015} measure the hand's ability to maintain a grasp in the presence of external disturbances. Resilience tests involve rapid disturbances simulating a collision with the environment, while grasp pullout tests involve static or quasistatic loading.

Some hand mobility benchmarks are based on anthropomorphic features. Feix et al.\ introduced a metric for evaluating the anthropomorphic motion capability of artificial hands, called the anthropomorphism index~\cite{feix_metric_2013}. This metric uses action manifolds, which describe all achievable fingertip poses for three-fingered grasps. The anthropomorphism index is computed from the overlap between a robot hand's action manifold and that of a human hand.

A well-known anthropomorphic benchmark of thumb mobility is the Kapandji test~\cite{kapandji_cotation_1986} (Figure \ref{fig:HandLevelReview}(b)), which originates in the occupational therapy literature. In this test, the thumb is scored in its ability to touch ten points on the fingers. 

Other thumb mobility benchmarks test the thumb's ability to work with fingers. The interactivity of fingers benchmark~\cite{you_kinematic_2019} is a weighted measure of the overlap between the workspace of each finger and the thumb. A related, more recent benchmark called KaRMA was proposed by Peticco and Agrawal to evaluate a hand's ability to translate and rotate a spherical object held in a two-finger pinch grasp~\cite{peticco_karma_2026} (Figure \ref{fig:HandLevelReview}(c)). From the KaRMA benchmark, three scores are produced at each point in the hand's workspace: KaRMA-T, KaRMA-R, and KaRMA-S. These scores measure translation ability, rotation ability, and sensitivity to grasp configuration, respectively.
For both the KaRMA and interactivity of fingers benchmarks, thumb and finger workspaces are discretized into voxels, simplifying computation. 

Typical measures of hand strength and speed are grasp strength and grasp cycle time~\cite{falco_grasping_2015}, tested, for example, using a split-cylinder artifact with embedded force sensors~\cite{falco_benchmarking_2020} (Figure \ref{fig:HandLevelReview}(d)). The grasp strength test evaluates the hand's maximum grasping strength with a wrap grasp and a pinch grasp~\cite{falco_grasping_2015, ma_yale_2017}. Grasp cycle time measures the time required for a hand to move from a stable grasp to the fully open configuration, then back to a stable grasp.

Hand dynamic properties can be characterized in terms of the dynamics of objects grasped by the hand. The spatial grasp compliance of a robot hand has been measured using an instrumented object mounted to the spindle of a CNC mill~\cite{odhner_compliant_2014}. The grasp impedance of human hands has been measured, including damping and inertial properties~\cite{burstrom_measurements_1990, hasser_system_2002, hoppner_grasp_2011}; however, a similar characterization has not been done for robot hands.

\subsection{Finger Benchmarks}
\label{subsection:Fingers}
Benchmarks of robot finger performance primarily focus on effort generation (force or torque), motion generation, and finger dynamics. Effort and motion-generation capabilities are based on maximum performance, accuracy, and precision, while finger dynamics measure controller response and the finger's dynamic properties. 

Fingertip strength is a common measure of a finger's force-generating capability. This is typically evaluated by measuring the maximum force a fully extended finger can generate in flexion at the fingertip~\cite{falco_benchmarking_2020}. The fully extended configuration is often a finger's weakest, so the maximum strength measured in this configuration can be achieved at any other configuration.

The fingertip force tracking test, as presented by NIST~\cite{falco_performance_2018}, measures force control and sensing accuracy at the fingertip. Fingers are tested with the fingertip pressed against a static load cell while a sinusoidal force profile is commanded at the fingertip, and external and internal force sensors are read out. Two measures are computed from the collected data. The first is the root-mean-square error (RMSE) between the commanded fingertip force and the externally measured fingertip force. This is a measure of the force control accuracy. The second measure is the RMSE between the externally and internally measured fingertip force. This indicates fingertip force sensing accuracy, which strongly influences fingertip force control.

With respect to motion generation, the finger's maximum speed is commonly reported in one of three ways: i) fingertip speed~\cite{bridgwater_robonaut_2012}, ii) joint speed~\cite{ShadowDEXEE, zhou_design_2024}, and iii) full range frequency~\cite{ShadowDexterousHand, roboteraxhand}. Fingertip speed simply measures the maximum speed the fingertip can achieve during an opening or closing motion; joint speed measures the maximum angular velocity achievable at the joints; and full range frequency, reported in~Hz, is the number of close-open cycles a finger can complete in one second.

Repeatability, a common measure in robotics, quantifies how consistently a finger can achieve a commanded configuration. A procedure for measuring finger repeatability was presented by NIST~\cite{falco_performance_2018, falco_benchmarking_2020}, in which a finger starts in a specified configuration, moves along a trajectory that requires all joints to move, and then returns to the initial configuration. Repeatability can be reported in joint space or fingertip task space as the error between the finger's initial and final configurations. 

Similarly, trajectory deviation is commonly used to quantify the error between a commanded trajectory and the true trajectory~\cite{shirafuji_development_2014, mnyusiwalla_new_2016, joshua_a_four-tendon_2021}, most often measured externally via computer vision. 

Finger dynamics are characterized by measuring the finger's response to control or environmental inputs. To characterize a finger's position control performance in free space, joint position step~\cite{abdallah_applied_2010} and frequency~\cite{jacobsen_design_1986} responses have been used. Similarly, joint torque control has been evaluated using both the step response~\cite{salisbury_articulated_1982} and the ramp response~\cite{abdallah_applied_2010}. 

Rao et al.\ analyzed the control bounds of robot finger compliance control~\cite{rao_analyzing_2017} using the joint-space actuator impulse response. In data sheets for robotic hands, it is increasingly common to list the backdriving torque of finger joints~\cite{roboteraxhand}. 
Otherwise, finger response to external excitations has received little attention.

\subsection{Wrist Benchmarks}
A well-designed wrist should enhance the mobility and stability of dexterous tasks while providing sufficient strength and shock protection for tasks requiring large forces and impacts~\cite{peticcoulloamarangola2025arxiv}.  Wrist designs can be categorized by whether they use serial~\cite{TownsendSalisbury1993Book, peticcoulloamarangola2025arxiv, BajajDollar2018ICBRB} or parallel~\cite{bridgwater_robonaut_2012, friedlhoppnerpetit2011iros, kimkimjang2018iros} mechanisms. If the hand is extrinsically actuated, load transmission through the wrist must also be considered~\cite{fanweiren2022robotica, bajajspiersdollar2019tro}.

Fan et al.\ developed a set of benchmarks for robot wrists that compare their capabilities to those of the human wrist~\cite{fanweiren2022robotica}. The set includes mobility, stability, maximum speed and torque, load capacity, and flexibility, each scored on a scale of 1 to 5. 
As noted in Section~\ref{sec:Hand}, mobility benchmarks include the number of wrist DOFs and range of motion~\cite{bajajspiersdollar2019tro}. Wrist stability measures the wrist's ability to maintain a configuration under external loading, quantified as the deviation angle from the desired pose.
Output capability measures the wrist's maximum torque and angular speed capabilities about each axis of rotation~\cite{peticcoulloamarangola2025arxiv, friedlhoppnerpetit2011iros, bajajspiersdollar2019tro, bridgwater_robonaut_2012}. Load capacity characterizes the structural failure of the wrist as a function of the load-to-weight ratio under bending loads~\cite{peticcoulloamarangola2025arxiv, friedlhoppnerpetit2011iros, fanweiren2022robotica, bajajspiersdollar2019tro}.

Finally, mechanical responsiveness to the environment is a desirable property for the wrist, and it has been primarily quantified by active~\cite{peticcoulloamarangola2025arxiv, friedlhoppnerpetit2011iros, kimkimjang2018iros} and passive~\cite{fanweiren2022robotica} backdriving torque.

\begin{table}[!ht]
\centering

\small
\setlength{\tabcolsep}{4pt}
\renewcommand{\arraystretch}{1.15}
\hspace*{\dimexpr(\textwidth-\typewidth)/2\relax}
\begin{tabularx}{\typewidth}{@{}
  l
  >{\raggedright\arraybackslash\hsize=1.35\hsize}X  % Hand metric
  >{\raggedright\arraybackslash\hsize=0.75\hsize}X  % Hand reference
  >{\raggedright\arraybackslash\hsize=1.35\hsize}X  % Finger metric
  >{\raggedright\arraybackslash\hsize=0.75\hsize}X  % Finger reference
  >{\raggedright\arraybackslash\hsize=0.90\hsize}X  % Wrist metric
  >{\raggedright\arraybackslash\hsize=0.90\hsize}X  % Wrist reference
  @{}}
\toprule
 & \multicolumn{2}{c}{\textbf{Hand}}
 & \multicolumn{2}{c}{\textbf{Finger}}
 & \multicolumn{2}{c}{\textbf{Wrist}} \\
\cmidrule(lr){2-3} \cmidrule(lr){4-5} \cmidrule(lr){6-7}
 & Metric & Reference & Metric & Reference & Metric & Reference \\
\midrule
\multirow{3}{*}{\rotatebox[origin=c]{90}{\textbf{Effort}}}
  & Pullout strength & \cite{kragten_proposal_2010,falco_grasping_2015}
  & Fingertip force & \cite{falco_benchmarking_2020}
  & Maximum torque & \cite{fanweiren2022robotica,peticcoulloamarangola2025arxiv,friedlhoppnerpetit2011iros,bajajspiersdollar2019tro,bridgwater_robonaut_2012} \\
  & Grasp strength & \cite{falco_grasping_2015,ma_yale_2017}
  & Fingertip force control accuracy & \cite{falco_performance_2018}
  & & \\
  & & & Fingertip force sensing accuracy & \cite{falco_performance_2018}
  & & \\
\midrule
\multirow{8}{*}{\rotatebox[origin=c]{90}{\textbf{Flow}}}
  & Mobility and connectivity & \cite{salisbury_articulated_1982}
  & Finger speed & \cite{bridgwater_robonaut_2012,ShadowDEXEE,zhou_design_2024,ShadowDexterousHand,roboteraxhand}
  & Maximum speed & \cite{fanweiren2022robotica,peticcoulloamarangola2025arxiv,friedlhoppnerpetit2011iros,bajajspiersdollar2019tro,bridgwater_robonaut_2012} \\
  & Anthropomorphism, anthropomorphic grasping ability & \cite{Feix2016TheGT,Cutkosky1989OnGC,llop-harillo_anthropomorphic_2019,feix_metric_2013}
  & Motion repeatability & \cite{falco_performance_2018,falco_benchmarking_2020}
  & Range of motion & \cite{bajajspiersdollar2019tro} \\
  & Grasp volume & \cite{kragten_proposal_2010,falco_grasping_2015}
  & Trajectory tracking & \cite{shirafuji_development_2014,mnyusiwalla_new_2016,joshua_a_four-tendon_2021}
  & & \\
%  & Anthropomorphism & \cite{feix_metric_2013} & & & & \\
  & Kapandji score & \cite{kapandji_cotation_1986} & & & & \\
  & Pinch grasp rotation and translation ability & \cite{peticco_karma_2026, you_kinematic_2019} & & & & \\
  & Grasp cycle time & \cite{falco_grasping_2015} & & & & \\
\midrule
\multirow{4}{*}{\rotatebox[origin=c]{90}{\textbf{Dynamics}}}
  & Grasp compliance & \cite{odhner_compliant_2014}
  & Step response, joint position and torque & \cite{salisbury_articulated_1982,abdallah_applied_2010}
  & Stability & \cite{fanweiren2022robotica} \\
  & & & Position frequency response & \cite{jacobsen_design_1986}
  & Flexibility & \cite{fanweiren2022robotica} \\
  & & & Compliance control bounds & \cite{rao_analyzing_2017}
  & Backdrive torque & \cite{fanweiren2022robotica,peticcoulloamarangola2025arxiv,friedlhoppnerpetit2011iros,kimkimjang2018iros} \\
  & & & Backdrive torque & \cite{roboteraxhand}
  & & \\
\midrule
\multirow{2}{*}{\rotatebox[origin=c]{90}{\textbf{Durability}}}
  & Impact resilience & \cite{negrello_benchmarking_2020,grebenstein_dlr_2011}
  & &
  & Load capacity & \cite{peticcoulloamarangola2025arxiv,friedlhoppnerpetit2011iros,fanweiren2022robotica,bajajspiersdollar2019tro} \\
  & Repeatability & \cite{christoph_orca_2025,zorin_ruka_2025,shaw_leap_2023}
  & & & & \\
  & Grasp resilience & \cite{negrello_benchmarking_2020,grebenstein_dlr_2011}
  & & & & \\ \\
\bottomrule
\end{tabularx}
\vspace*{0.2in}
\caption{A summary of existing hand-level benchmark metrics grouped by effort, flow, dynamics, and durability. Hand grasping metrics are grouped with the motion benchmarks.}
\label{tab:benchmark-metrics}

\end{table}

\subsection{Perspective}
\subsubsection{Gaps in benchmarking}
Current hand-level benchmarks (Table~\ref{tab:benchmark-metrics}) evaluate the hand's force production or motion capabilities individually, while dynamic performance is rarely measured. This can be seen in robot hand datasheets, where metrics follow those used for industrial robotic arms, which are generally designed for precise positioning in free space, not sensitive interaction with objects and the environment.

\subsubsection{Addressing the gaps}
\label{sec:HandLevelPerspective}
Dynamic interactions can be controlled using impedance control or admittance control.
Impedance control closes motion feedback loops around a force-controlled plant, while admittance control closes force feedback loops around a motion-controlled plant~\cite{hogan_impedance_1984}.

Because stability and passivity impose fundamental limits~\cite{colgate_analysis_1989}, feedback control cannot arbitrarily reshape plant dynamics. Therefore, each approach is best suited to a different regime.
Impedance control is preferable when the desired impedance is low (close to the native force-controlled plant), while admittance control is preferable when the desired impedance is high (close to the native motion-controlled plant).
Robot arms, which are large and primarily used for motion control, have benefited most from admittance control.
Robot fingers, however, are small and frequently used to regulate contact forces, for example, when pressing a button, mating a connector, or seating a lid on a jar, making impedance control a particularly promising approach.

Good impedance control requires \textit{mechanical transparency}: forces from the actuators should be faithfully conveyed to the environment, and forces imposed by the environment should be faithfully conveyed back to the actuator, where they can be sensed and controlled.
This bidirectional fidelity is sometimes called \emph{bilateral force control}~\cite{Lawrence1992StabilityAT}.
To understand why transparency is difficult to achieve, consider the typical actuation chain: a motor drives a reduction mechanism (gearbox, linkages), and a remotization element (cables, pulleys, joints) carries force and motion to the fingertip (Figure~\ref{fig:FingerDynamics}(a)).
Stages in this chain introduce compliance, inertia, and nonlinear effects such as friction, backlash, cogging, and torque ripple.
These factors distort the transmission of forces and motions in both directions.

We propose three metrics to characterize mechanical transparency, and each can be assessed in both the \textit{forward} direction (actuator to fingertip) and the \textit{backward} direction (fingertip to actuator):

\begin{figure}
    \centering
    
    \includegraphics[width=\textwidth]{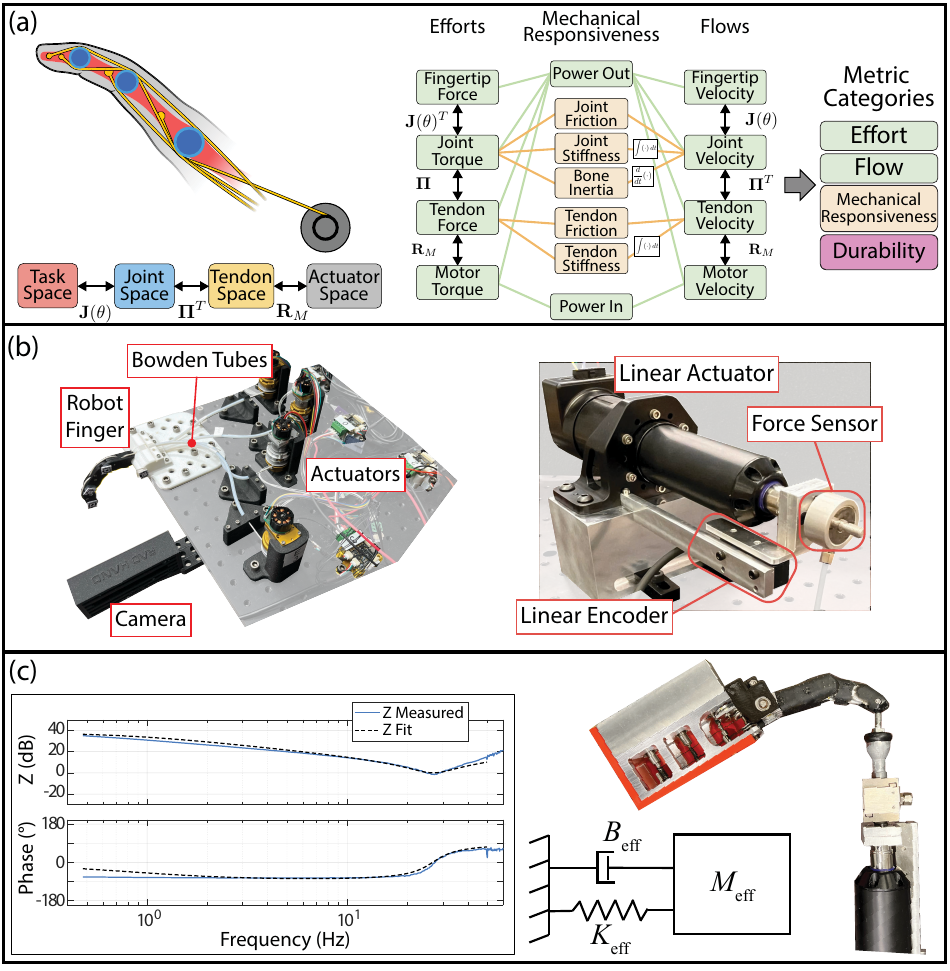}

    \caption{
    (a) The mapping of efforts and flows between actuators and fingers describes hand transparency. A generalized physical model provides categories for benchmarking hand-level physical performance based on efforts, flows, and mechanical responsiveness. The durability category is added, as it is critical to real-world deployment of hands.
    (b) HAND finger testbeds under development at Texas A\&M University (left) and Northwestern University (right). The non-backdrivable fingertip dynamometer (right) with collocated force and position sensing at the output is used to measure finger dynamic properties.
    (c) The mechanical impedance frequency response for the unpowered robot finger~\cite{batteas2025design} as measured by the fingertip dynamometer in (b).
    A dynamic model is fit to the data, and the effective mass, stiffness, and damping of the finger is found to be $M_\mathrm{eff} = 13.6$~g, $K_\mathrm{eff} = 389$~N/m, and $B_\mathrm{eff} = 1.0$~Ns/m. For reference, a human finger measured with the same device yielded values of $M_\mathrm{eff} = 7.0$~g, $K_\mathrm{eff} = 185$~N/m, and $B_\mathrm{eff} = 1.02$~Ns/m.
    }
    
    \label{fig:FingerDynamics}
    
\end{figure}

\begin{enumerate}

    \item \textbf{Breakaway force} is the minimum force required to overcome static friction in the transmission.
    In the forward direction, it is the smallest commanded actuator effort that produces fingertip motion, which can be measured by commanding a ramp of fingertip force while the finger is free to move.
    In the backward direction (backdrive breakaway force), it is the smallest force applied to the fingertip that produces actuator motion, which can be measured by applying a force ramp to the fingertip while recording actuator positions.
    Because reduction mechanisms amplify friction, backdrive breakaway force is typically much larger than forward breakaway force.
    Hands with non-backdrivable transmissions should not be tested for backdrive breakaway force, as doing so risks mechanical failure.

    \item \textbf{Force bandwidth} characterizes how rapidly the system responds to a changing force command or to a changing environmental displacement.
    Bandwidth is governed primarily by the mass and compliance that lie in the transmission path between actuators and the fingertip.
    In the forward direction, it may be measured by pressing the fingertip against a force sensor and commanding a sinusoidal frequency sweep of fingertip force; the bandwidth is the frequency at which the closed-loop response drops to $-3$\,dB.
    In the backward direction, bandwidth is related to the effective mechanical impedance of the fingertip.  Ideally, the impedance is low so that a displacement imposed by the environment does not cause an unduly large reaction force at the finger.
    A non-backdrivable dynamometer (Figure~\ref{fig:FingerDynamics}(b)) can impose a velocity sine-sweep excitation on the fingertip while measuring the interaction force; effective mass, stiffness, and damping are then extracted by fitting a dynamic model to the impedance frequency response (Figure~\ref{fig:FingerDynamics}(c)).
    Note that these impedance measurements provide a linear approximation of the finger's local dynamics; in practice, fingertip impedance varies with configuration, and transmission nonlinearities produce amplitude- and history-dependent behavior that this model cannot capture.

    \item \textbf{Force accuracy} quantifies how closely the transmitted force tracks a desired profile.
    Accuracy is limited primarily by friction and the nonlinear factors described above, though feedback control can partially compensate for these effects.
    In the forward direction, accuracy may be measured using the NIST finger force tracking benchmark~\cite{falco_benchmarking_2020}: the fingertip is pressed statically against a force sensor, a sinusoidal force profile is commanded, and the RMSE between commanded and measured force is reported.
    %In the backward direction, a sinusoidal force can be imposed on the fingertip and the resulting force or torque at the actuators measured.

\end{enumerate}

Designing for transparency is a promising approach to enabling high-quality impedance control for dynamic tasks. Mechanical transparency also increases system durability by reducing impact forces from unexpected contact with the environment and by reducing internal wear through low-friction designs.
However, many other hand-level performance measures significantly impact system-level dexterity.
Benchmarks for speed, strength, and mobility that we expect to appear in future hand datasheets are described on the HAND ERC benchmarking website.

% Section 4
\section{COMPONENT-LEVEL BENCHMARKING}

\label{sec:LowLevel}
\subsection{Actuation Benchmarks}
Actuators generate motion and force in robot hands. As highlighted in Section~\ref{sec:HandLevelPerspective}, excellent force regulation is a key enabler of dexterity. Therefore, the ideal dexterous actuator is a perfect force source, enabling the exact forces commanded by a controller to be imposed on the environment. 

Electromagnetic (EM) motors are the most common type of robot hand actuator due to their maturity, torque density, and reliability~\cite{Saliba2013-hg_A16, lin_comprehensive_2025}.
Other actuators used in robotic hands include rigid pneumatic cylinders~\cite{jacobsen_design_1986} and soft pneumatic chambers~\cite{puhlmann2022rbo}. Although less mature, soft actuators, such as hydraulically amplified self-healing electrostatic actuators (HASELs), liquid crystal elastomers (LCEs), and dielectric elastomer actuators (DEAs), seek to replicate the performance of human muscle~\cite{Mirvakili2018-rv_A14}.

Significant work has been done to establish a framework for evaluating actuator performance across different technologies. Early work by Hollerbach et al.\ surveyed many classes of actuators, their mechanisms, and robotics applications, using metrics normalized by mass such as specific power (power per unit mass) and specific force/torque (force or torque per unit mass)~\cite{Hollerbach1992-nk_A12}. However, uneven treatment of which components are included in the mass accounting (such as power electronics) can make accurate comparisons difficult. Given the strict size constraints of humanoid hands, the same metrics normalized by volume are also of interest~\cite{Saliba2013-hg_A16}. In the case of linear actuators, stroke length is a critical metric (as all control must lie within this limited range) and is sometimes presented as strain~\cite{Hollerbach1992-nk_A12, Mirvakili2018-rv_A14, Saliba2013-hg_A16, Greco2022-gr_A10}. Maximum force (or torque) is similarly important~\cite{Rupert2021-ea_A9}. However, these static metrics fail to describe the system's dynamic performance, which is of particular importance given the nature of many dexterous tasks.

Typical metrics for dynamic actuator performance include step response characteristics, such as rise time or overshoot, as well as control bandwidth~\cite{Towards_ACT_Char2026}. Actuator dynamometers (Figure \ref{fig:ComponentReview}(a)) can be used to estimate control bandwidth and evaluate performance under various loading conditions~\cite{Towards_ACT_Char2026}. Sakama et al.\ complemented the static normalized performance measures with additional ones aimed at responsiveness, preferring power rate (torque squared per inertia) and specific power rate (power rate per mass) over torque per inertia, which changes based on the transmission~\cite{Sakama2022-dl_A11}. Force-velocity and speed-torque curves are also important, particularly in artificial muscles which exhibit a strong relationship between strain rate and force production~\cite{Hollerbach1992-nk_A12}. 

Saliba et al.\ highlight linearity, smoothness, and hysteresis as qualitative metrics to evaluate actuators for dexterity~\cite{Saliba2013-hg_A16}. Force nonlinearities due to stroke limits, hysteresis in soft actuators, and backlash in a gearbox can significantly affect an actuator's ability to dynamically regulate its output force. For soft actuators, Rupert et al.\ discuss parasitic and variable stiffness to capture torque loss and stiffness changes, respectively~\cite{Rupert2021-ea_A9}. Parasitic stiffness occurs when compliant structural elements push or pull against the actuator. Fingers that use a spring for extension experience parasitic stiffness~\cite{you_design_2019}.

\subsection{Mechanical Transmission Benchmarks}
\label{subsection:MechanicalTransmission}

A mechanical transmission transfers efforts and flows between the joints and actuators of a hand (Figure~\ref{fig:FingerDynamics}(a)). As the primary element between the environment and the actuators, the characteristics of a transmission have a large influence on the transparency of the system. In intrinsically-actuated robotic hands, the mechanical transmission elements may be small linkages~\cite{kim_integrated_2021} or gearboxes. In extrinsically actuated robotic hands, the mechanical transmission becomes a more significant and independently designable element, often crossing multiple degrees of freedom to deliver power using elements such as tendons~\cite{NazmaMohd2012JMERR} or fluids~\cite{SchwarmGravesmillWhitney2019ICRA}.

The most significant parameter related to transmissions is mechanical advantage, often presented as the ratio of the input force (or torque) to the output force (or torque). In the case of  linkages, tendons, and soft transmissions, the mechanical advantage may be configuration dependent. Mechanical advantages fundamentally distort the relationship between the forward (actuator to environment) and backward (environment to actuator) transparencies. In the context of robotic hands, large mechanical advantages are often necessary due to limited space and mass budget for high-torque actuators. Friction, which occurs in the transmission elements~\cite{ PalliBorghesanMelchiorri2009ICRA, ReineckeChalonFriedl2014ICRA, Gogoussis_Friction_1988ICRA} and at the joints themselves~\cite{borghesan_design_2010}, adds significantly to this distortion. Frictional losses are sometimes described as a power transfer efficiency ratio~\cite{SeokWangChuah2014TMECH}. In addition to friction, other nonlinearities, such as backlash in gears or hysteresis in Bowden cables, adversely affect transparency~\cite{Saliba2013-hg_A16}. Other effects relevant to robot hands include the coupling of DOFs that the transmission crosses~\cite{jacobsen_design_1986, TakeiFrishmanWhitney2025ICRA} and hardware complexity~\cite{kim_integrated_2021}.

Beyond static force transmission, the dynamic properties of a transmission strongly influence hand control. High compliance from tendons (in comparison to metal gears and linkages) can reduce position-control bandwidth~\cite{KanekoYamashitaTanie1991ICAR}, while large reflected inertia from mechanical advantage raises the impedance displayed by the finger. 

\begin{figure}
    \includegraphics[width=\textwidth]{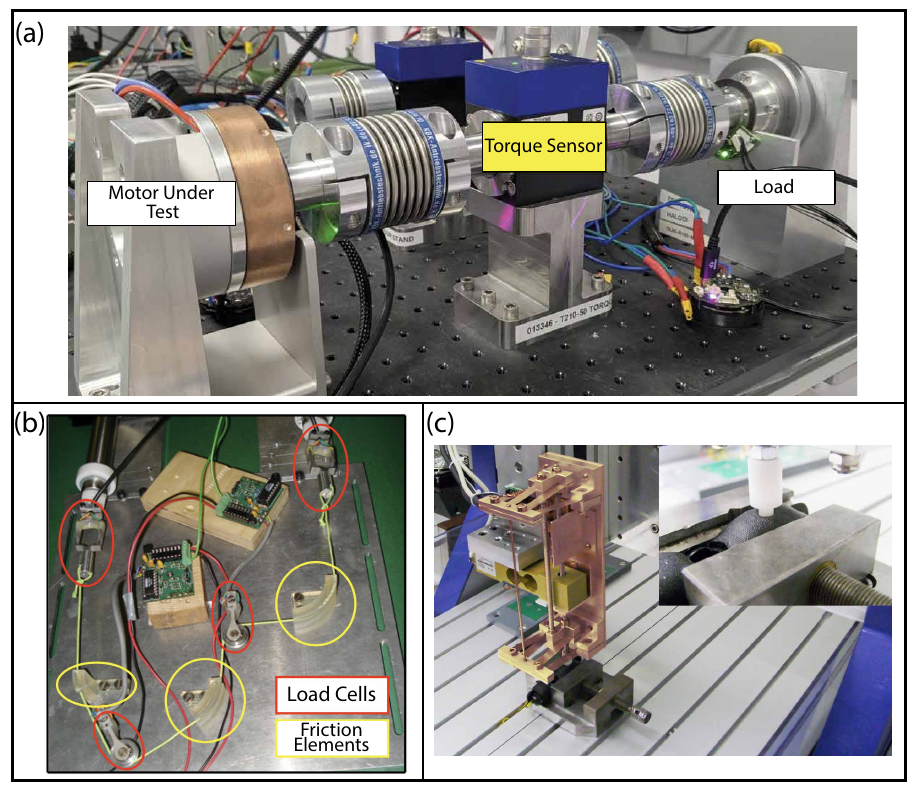}
    \caption{Existing component-level benchmarking testbeds. 
    (a) An actuator dynamometer designed to evaluate the dynamic performance of electric motors (adapted from \cite{Towards_ACT_Char2026} with permission).
    (b) Routing elements and load cells are used to characterize frictional losses in a tendon transmission with sliding surfaces (adapted from~\cite{PalliBorghesanMelchiorri2009ICRA} with permission).
    (c) A multi-axis stage equipped with a force sensor is used to measure the response of a tactile sensor (adapted from~\cite{Buscher2015IROS} with permission).
    }
    \label{fig:ComponentReview}
\end{figure}

Given the prevalence of tendons in robotic hands due to their compact and lightweight nature~\cite{Lovchik1999, grebenstein_dlr_2011, jacobsen_design_1986, liu_multisensory_2008, bridgwater_robonaut_2012, salisbury_articulated_1982}, benchmarks have been developed to characterize their behavior and performance. A tendon transmission consists of three primary elements: the tendon itself, routing elements that guide the tendon between the actuator and driven joint, and terminations that connect the tendon to the surrounding structure.
Tendons are typically metal cables~\cite{Lovchik1999} or synthetic fiber cords~\cite{grebenstein_dlr_2011, TasiPelyva2025ICORR, LiuZhengHliboky2024Biomimetics}. Common benchmarks for tendon materials evaluate maximum tensile strength~\cite{AsaneSchmitz2020IROS, HorigomeEndoTakata2018IEEERAL, MazumdarSpencerHobart2017IEEERAL}, lifetime under various loading conditions~\cite{AsaneSchmitz2020IROS, MazumdarSpencerHobart2017IEEERAL, HorigomeEndo2018IEEERAL}, stiffness, elongation over lifetime (creep), and strength degradation.
Routing elements used to guide the tendon include pulleys supported by bearings or bushings, flexible Bowden cables, and rigid sliding surfaces~\cite{jacobsen_utahmit_1984,bridgwater_robonaut_2012, salisbury_articulated_1982}. These routing elements are often evaluated for tendon-path friction~\cite{PalliBorghesanMelchiorri2012TRO, PalliBorghesanMelchiorri2009ICRA, ReineckeChalonFriedl2014ICRA} and hysteresis (Figure \ref{fig:ComponentReview}(b)). Tendon life-cycle tests are typically performed using the expected routing elements, as friction adversely impacts the lifespan. 

Termination methods vary by tendon material and include crimps, potting, knots, and splicing~\cite{HorigomeEndoTakata2018IEEERAL, grebenstein_antagonistically_2010}. Termination strength is commonly measured alongside tendon material strength because the termination is frequently the limiting failure point~\cite{HorigomeEndoTakata2018IEEERAL, MazumdarSpencerHobart2017IEEERAL}. Despite their prevalence, tendons are a primary source of failure in many robotic hands, underscoring the need to understand how materials, routing elements, and terminations affect performance and lifespan.

\subsection{Soft, Durable Skin Benchmarks}
The skin of a robot hand endures repeated contact and abrasion while shielding the components underneath. In some cases, the skin covers a tactile sensing layer, and its thickness, modulus, and surface texture determine how contact forces are distributed to the sensors. If the skin is soft, it can yield large contact patches at low forces, stabilize grasps against disturbances, and simplify the control of roll-slide contacts. A skin's softness and its durability tend to be inversely correlated, however, and as a result, many robot hand manufacturers avoid using soft skins. 

When robot hands are equipped with skins, they are often formed from rubber, elastomers such as polydimethylsiloxane (PDMS)~\cite{Shintake2018, Porte2024}, platinum-cure silicones~\cite{Porte2024}, thermoplastic polyurethane (TPU)~\cite{Georgopoulou2021}, and foams~\cite{Bauer2020}. These materials span a wide range of mechanical behavior and are typically characterized by metrics such as Young’s modulus, tensile strength, elongation at break, and Shore hardness~\cite{Shintake2018}. Skin compliance increases the area of contact patches, but viscoelastic properties can cause surface adhesion that produces stick-slip sliding that adversely impact dexterous manipulation~\cite{Shintake2018, Porte2024}.

Skin friction enables in-hand manipulation and fine motor tasks. The coefficient of friction, the maximum tangential force normalized by the applied normal load, is most often measured by reciprocating sliding against a defined counter surface or estimated using contact mechanics modeling~\cite{Herr2023, Roels2026}. The friction of elastomers can be tuned by altering material stiffness by modifying chemistry, applying coatings, or patterning the surface topography. Hao et al.\ measured the effect of skin texture on friction, with the effect depending on pattern and lubrication~\cite{Hao2024} (Figure \ref{fig:ComponentReview}(c)). Other work reports stiffness or contact area without friction~\cite{Porte2024,Georgopoulou2021,Berthold2023,Romero2020}. 

Wear mechanisms for these elastomers are not well studied, and durability is reported only qualitatively. 
 
\subsection{Sensing Benchmarks}
A robot hand’s ability to interact with the world depends on two fundamental sensing capabilities: sensing the configuration and joint loading conditions (i.e., kinesthetic sensing) and sensing the physical contact it makes with the environment (i.e., tactile sensing). The field has converged on several key dimensions for benchmarking performance of both: operating range describes the span of values a sensor can measure; resolution captures the smallest detectable change; accuracy quantifies deviation from ground truth; and sensing frequency determines the rate of data acquisition.

Benchmarking of kinesthetic sensing has received relatively little attention in the literature. Joint positions, velocities, and accelerations are typically measured using optical~\cite{Yu2021}, capacitive~\cite{Pu2018}, or magnetic~\cite{Paredes2021, jacobsen_design_1986, bridgwater_robonaut_2012, ShadowDexterousHand}  encoders. For commercially available sensors, most performance metrics can be directly sourced from manufacturers’ datasheets. When custom implementations are evaluated, external cameras are used to track fiducials as the ground truth for joint positions. Sensing joint loading typically relies on strain gauges~\cite{Lovchik1999}, series-elastic actuators~\cite{PrattWilliamson1995ICRA, Koochakzadeh2024}, or proprioceptive actuation~\cite{Wensing2017}, and each can be benchmarked against known loads or external measurements. The most notable benchmarking effort is from NIST~\cite{falco_grasping_2015}, which evaluates fingertip force measurement accuracy by computing the error between the internal force-sensing and the external force-sensor reading. However, benchmarking individual sensor capabilities in isolation is insufficient; fitting sensors within the confined geometry of a robot hand is a critical practical constraint. Metrics that incorporate volumetric considerations, such as volume-specific resolution (resolution per unit sensor volume), are needed to fully capture the tradeoff between kinesthetic sensor performance and the space it consumes.

Benchmarking of tactile sensing has received considerably more attention and remains an active area of development. Tactile sensors transduce mechanical interactions, such as contact location and force distribution, into electrical signals through piezoresistive~\cite{13Kim2024, Park2022, Buscher2015IROS, Xu2023} (Figure \ref{fig:ComponentReview}(c)), piezoelectric~\cite{Tang2025, Zhang2022}, capacitive~\cite{Cheng2023, Sarwar2023, Su2021}, magnetic~\cite{Hu2024, Pattabiraman2025, Bhirangi2021}, triboelectric~\cite{Liu2025_Triboelectric, Song2022}, and optical~\cite{Yuan2017, Lambeta2024, Zhao2025, Do2022} methods; in-depth reviews of each modality can be found in~\cite{Jassim2025, Lepora2026, Meribout2024}. 

In robotics, optical-based tactile sensors have received widespread attention due to their reliance on well-developed vision technologies, adaptability to different robot hands, and ease of integration with machine learning protocols~\cite{Li2025}. Cong et al.~\cite{Cong2026} proposed many benchmarks for vision-based tactile sensors, including camera resolution, field of view, sensor gel thickness, frame rate, calibration error, spatial resolution, sensitivity, repeatability, spatial robustness, and lighting robustness. While intended for optical sensors, many of these metrics extend to other modalities. Complementing this, Liu and Ward-Cherrier~\cite{Liu2024_Haptics} proposed a sensor-agnostic Tactile Image-Based Psychophysics-Inspired benchmark, which evaluates spatial acuity across accuracy, stability, and generalizability. Notably, this framework not only characterizes sensor performance but also guides hardware optimization. 

Friedl and Roa~\cite{Friedl2021} benchmarked nine tactile sensors on a variable-stiffness gripper, finding that gripper stiffness and actuator angular velocity significantly affect sensor performance. This highlights that element-level metrics alone may not capture real-world behavior under integration conditions.

\begingroup
\setlength{\LTleft}{\fill}
\setlength{\LTright}{\fill}
\small
\setlength{\tabcolsep}{4pt}
\renewcommand{\arraystretch}{1.25}
\begin{xltabular}{\typewidth}{@{}
  l
  >{\raggedright\arraybackslash\hspace{0pt}\hsize=1.10\hsize}X  % Metric
  >{\raggedright\arraybackslash\hspace{0pt}\hsize=0.8\hsize}X  % Reference
  @{}}

\toprule
 & Metric & Reference \\
\midrule
\endfirsthead
\multicolumn{3}{@{}l}{\footnotesize\itshape Table \thetable\ (continued)} \\
\toprule
 & Metric & Reference \\
\midrule
\endhead
\midrule
\multicolumn{3}{@{}r}{\footnotesize\itshape continued on next page} \\
\endfoot
\bottomrule
\caption{Existing component-level metrics, grouped
by subsystem.}
\label{tab:component-metrics} 
\endlastfoot

\textbf{Actuators}
  & Volume & \cite{Saliba2013-hg_A16,liu_multisensory_2008,jacobsen_design_1986} \\*
  & Density & \cite{ Greco2022-gr_A10} \\*
  & Mass & \cite{Saliba2013-hg_A16,liu_multisensory_2008,jacobsen_design_1986} \\*
  & Complexity & \cite{Saliba2013-hg_A16} \\*
  & Specific power (power/mass) & \cite{Hollerbach1992-nk_A12,Greco2022-gr_A10,Sakama2022-dl_A11,Saliba2013-hg_A16} \\*
  & Specific force, torque (torque/mass) & \cite{Hollerbach1992-nk_A12,Rupert2021-ea_A9} \\*
  & Stroke length, strain & \cite{Mirvakili2018-rv_A14,Saliba2013-hg_A16,Hollerbach1992-nk_A12,Greco2022-gr_A10,Rupert2021-ea_A9,bridgwater_robonaut_2012} \\*
  & Maximum force, torque & \cite{Rupert2021-ea_A9,Sakama2022-dl_A11,bridgwater_robonaut_2012,liu_multisensory_2008,jacobsen_design_1986} \\*
  & Parasitic and variable stiffness & \cite{Rupert2021-ea_A9} \\*
  & Power rate (torque$^2$/inertia) & \cite{Sakama2022-dl_A11} \\*
  & Control bandwidth & \cite{Hollerbach1992-nk_A12,Towards_ACT_Char2026,jacobsen_design_1986} \\*
  & Speed-torque behavior & \cite{Hollerbach1992-nk_A12} \\*
  & Linearity, hysteresis, and force smoothness & \cite{Saliba2013-hg_A16} \\*
  & Efficiency & \cite{Rupert2021-ea_A9,Greco2022-gr_A10,Saliba2013-hg_A16} \\
\addlinespace[2pt]
\midrule
\shortstack{\textbf{Transmission}\\(general)}
  & Hardware complexity & \cite{kim_integrated_2021,Saliba2013-hg_A16} \\*
  & Transmission ratio & \cite{liu_multisensory_2008} \\*
  & Compliance and position-control bandwidth & \cite{KanekoYamashitaTanie1991ICAR} \\*
  & Coupling & \cite{jacobsen_design_1986,TakeiFrishmanWhitney2025ICRA,Saliba2013-hg_A16} \\*
  & Hysteresis, backlash, and inertia & \cite{Saliba2013-hg_A16} \\*
  & Transmission friction (efficiency) & \cite{PalliBorghesanMelchiorri2009ICRA,ReineckeChalonFriedl2014ICRA,Gogoussis_Friction_1988ICRA,borghesan_design_2010,SeokWangChuah2014TMECH} \\
\addlinespace[2pt]
\midrule
\shortstack{\textbf{Transmission}\\(tendons)}
  & Tendon diameter & \cite{bridgwater_robonaut_2012,christoph_orca_2025} \\*
  & Routing friction & \cite{PalliBorghesanMelchiorri2012TRO, PalliBorghesanMelchiorri2009ICRA, ReineckeChalonFriedl2014ICRA} \\*
  & Termination strength & \cite{HorigomeEndoTakata2018IEEERAL, MazumdarSpencerHobart2017IEEERAL} \\*
  & Material strength & \cite{AsaneSchmitz2020IROS,HorigomeEndoTakata2018IEEERAL,MazumdarSpencerHobart2017IEEERAL,bridgwater_robonaut_2012} \\*
  & Hysteresis & \cite{PalliBorghesanMelchiorri2009ICRA} \\*
\addlinespace[2pt]
\midrule
\shortstack{\textbf{Sensors}\\(tactile)}
  & Thickness & \cite{Buscher2015IROS,Park2022,Xu2023,Tang2025,Cong2026,Cheng2023,Sarwar2023,Su2021,Hu2024,Bhirangi2021,Yuan2017,Lambeta2024,Do2022} \\*
  & Sensing field size & \cite{Cong2026,Tang2025,Zhang2022,Sarwar2023,Hu2024,Bhirangi2021,Yuan2017,Do2022,Jassim2025,Li2025} \\*
  & Coverage & \cite{Zhao2025} \\*
  & Temporal resolution & \cite{Cong2026,13Kim2024,Xu2023,Tang2025,Zhang2022,Su2021,Bhirangi2021,Liu2025_Triboelectric,Yuan2017,Lambeta2024,Do2022,Jassim2025,Lepora2026,Li2025,Friedl2021} \\*
  & Spatial resolution & \cite{Cong2026,Park2022,Xu2023,Liu2024_Haptics,Tang2025,Zhang2022,Sarwar2023,Hu2024,Pattabiraman2025,Bhirangi2021,Yuan2017,Lambeta2024,Zhao2025,Do2022,Jassim2025,Lepora2026,Li2025,Friedl2021} \\*
  & Sensing range & \cite{13Kim2024,Park2022,Xu2023,Tang2025,Zhang2022,Cheng2023,Sarwar2023,Su2021,Hu2024,Pattabiraman2025,Liu2025_Triboelectric,Jassim2025,Lepora2026,Friedl2021,bridgwater_robonaut_2012} \\*
  & Force sensitivity & \cite{Cong2026,13Kim2024,Park2022,Xu2023,Tang2025,Zhang2022,Cheng2023,Sarwar2023,Su2021,Liu2025_Triboelectric,Lambeta2024,Jassim2025,Lepora2026,Li2025,christoph_orca_2025,falco_grasping_2015} \\*
  & Noise, accuracy & \cite{Park2022,Sarwar2023,Pattabiraman2025,Bhirangi2021,Lambeta2024,Jassim2025,Lepora2026,Liu2024_Haptics,bridgwater_robonaut_2012,jacobsen_design_1986, Cong2026} \\*
  & Detection limit & \cite{13Kim2024,Park2022,Buscher2015IROS,Xu2023,Tang2025,Zhang2022,Cheng2023,Sarwar2023,Su2021,Pattabiraman2025,Bhirangi2021,Liu2025_Triboelectric,Yuan2017,Jassim2025,Lepora2026} \\*
  & Stretchability & \cite{Cheng2023,Su2021} \\*
  & Repeatability & \cite{Cong2026,13Kim2024,Buscher2015IROS,Park2022,Zhang2022,Cheng2023,Sarwar2023,Tang2025,Su2021,Hu2024,Jassim2025,Lepora2026} \\*
  & Calibration error & \cite{Cong2026,falco_grasping_2015} \\*
  & Durability & \cite{Park2022, Buscher2015IROS, Tang2025, Bhirangi2021, Song2022, Do2022, Jassim2025, Lepora2026} \\
\addlinespace[2pt]
\midrule
\shortstack{\textbf{Sensors}\\(kinesthetic)}
  & Volume & \cite{liu_multisensory_2008} \\*
  & Temporal resolution & \cite{Paredes2021,ShadowDEXEE,ShadowDexterousHand} \\*
  & Spatial resolution & \cite{Paredes2021,Yu2021,Pu2018,ShadowDEXEE,ShadowDexterousHand} \\*
  & Measurement range and sensitivity & \cite{Yu2021} \\*
  & Accuracy & \cite{Paredes2021,Yu2021,Pu2018,falco_grasping_2015} \\*
  & Durability & \cite{Pu2018} \\*
  & Robustness & \cite{Paredes2021} \\
\addlinespace[2pt]
\midrule
\textbf{Skin}
  & Stress-strain & \cite{Shintake2018,Porte2024,Georgopoulou2021,Herr2023,Roels2026,falco_grasping_2015} \\*
  & Relaxation percentage & \cite{Porte2024,Georgopoulou2021,Roels2026} \\*
  & Glass transition temperature, peak melting temperature, and curing time  & \cite{Roels2026} \\*
  & Coefficient of friction & \cite{Herr2023,Hao2024,Berthold2023} \\*
  & Wear and wear rate 
  & \cite{Herr2023} \\
\end{xltabular}
\endgroup

\subsection{Perspective}
\subsubsection{Gaps in benchmarking}
Existing benchmarks (Table~\ref{tab:component-metrics}) have largely been developed within individual component research communities, where well-established metrics are used to characterize and compare different technologies. However, these metrics are designed to assess component-level performance in isolation. When these components are integrated into a robot hand for dexterous manipulation, additional requirements, constraints, and tradeoffs emerge.  

For actuators and mechanical transmission, a key gap is the lack of standardized metrics for evaluating their combined contribution to the mechanical transparency of a robot hand. 
Transmission friction and efficiency metrics are not yet standardized, and benchmarking transmissions based on tendons is complicated by configuration dependence and the diversity of routing elements.

For tactile sensing, we currently lack standardized representations and metrics that allow direct comparison of different technologies, including their suitability to scaling to whole-hand coverage.

For soft skin, substantial attention has been devoted to characterizing material properties. However, less effort has been given to capturing the functional role of the skin in robot hands. The skin can provide stabilizing compliance and enable large, conformal contact areas, but there are no standardized metrics for reporting how contact area evolves during interaction. 

\subsubsection{Addressing the gaps}

\begin{figure}
    \includegraphics[width=\textwidth]{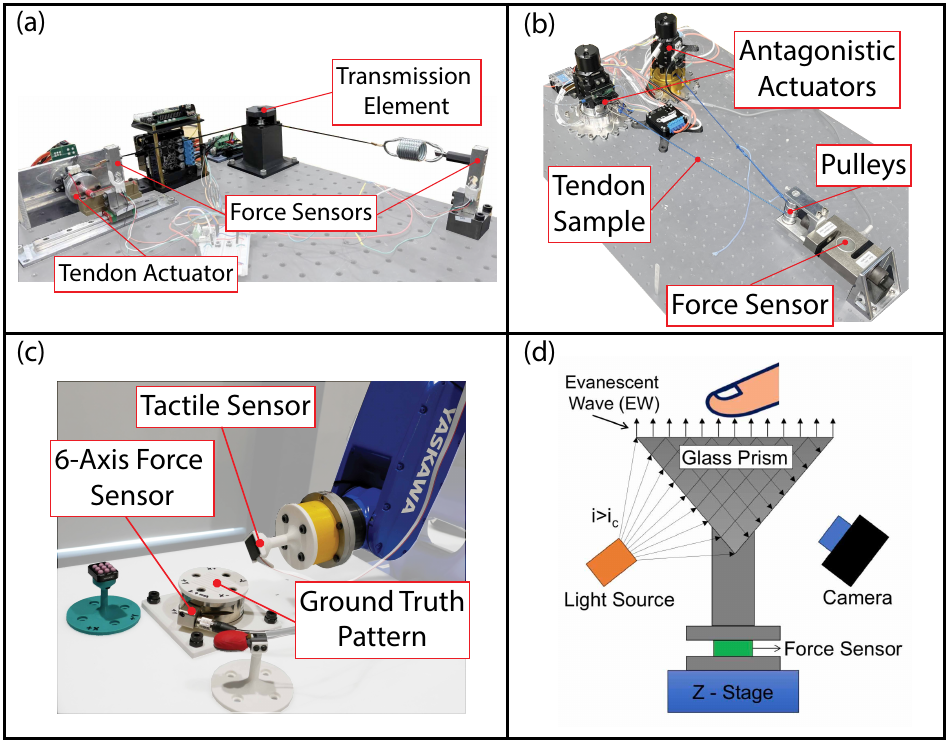}
    \caption{Component-level testbeds built by members of the HAND ERC. 
    (a) A tendon transmission testbed at Northwestern University that is used to evaluate the transmission efficiency of tendons and routing elements.
    (b) A tendon strength and lifetime testbed at Texas A\&M University with two antagonistic motors that pull a tendon over a sliding or rolling surface. 
    (c) A tactile sensing testbed at Carnegie Mellon University that is used to verify tactile sensing accuracy and resolution.
    (d) A contact patch testbed at Texas A\&M University that measures contact patch evolution for human and robot fingertips using frustrated total internal reflection (FTIR).
    }
    \label{fig:ComponentTestbeds}
\end{figure}

Addressing these gaps requires standardized methods for evaluating hand components, particularly in the context of dexterity. The HAND ERC is currently developing a set of component-level benchmarks and equipment to evaluate them. These benchmarks are evolving and are reported on the HAND benchmarking website. Figure~\ref{fig:ComponentTestbeds} depicts several component testbeds currently under development, including a tendon strength and lifetime testbed, to better understand common tendon failure conditions (Figure~\ref{fig:ComponentTestbeds}(b)); a testbed to evaluate tactile sensor spatial resolution and sensitivity (Figure~\ref{fig:ComponentTestbeds}(c)); and a testbed to measure contact area evolution under different loading conditions of human and robot skin (Figure~\ref{fig:ComponentTestbeds}(d)).

As described in Section~\ref{sec:HandLevelPerspective}, the hand's mechanical transparency is critical to sensitive controlled interactions with the environment. Three component-level performance measures strongly influence hand transparency:
\begin{enumerate}
    \item \textbf{Actuator output inertia} impacts force control bandwidth, with higher inertia reducing the finger's responsiveness to environmental interactions. A key design trade-off arises here: decreasing actuator inertia comes at the cost of reduced strength, while strength amplification via a transmission element increases inertia by the transmission ratio squared. Metrics such as specific torque and actuator torque per reflected inertia can be used to characterize this tradeoff.
    \item \textbf{Transmission friction} arises from sliding between surfaces along the hand's transmission path, which creates a discrepancy between input and output force. Because friction is nonlinear, it is approximated by the force transmission efficiency, the ratio of output force to input force. At small forces, motion through the transmission can be lost to stiction; this effect is captured by the transmission breakaway force, the minimum force applied at either end of the transmission that produces motion at the other end (Figure \ref{fig:ComponentTestbeds}(a)).
    \item \textbf{Transmission stiffness} plays a major role in the overall actuation/transmission impedance, and therefore it should be rigorously characterized.
    Transmission elements that have low stiffness, either because they were chosen for that purpose (e.g., series-elastic actuators) or as a consequence of their other purposes (e.g., tendons), decrease the apparent impedance of the finger to the environment but reduce the force control bandwidth.  

\end{enumerate}

% Section 5
\section{DISCUSSION}

Perhaps the most pressing research problem in robot hand design and benchmarking is understanding how low-level design decisions impact benchmark performance at the system and application level. In this paper, we propose component- and hand-level benchmarks which, based on our collective experience in robot manipulation, design, and control, will help bridge the attribution gap between low-level design decisions and system-level performance. The HAND ERC is currently developing test equipment to perform these component- and hand-level benchmarks, as described above.

Still, the hard work of establishing system performance sensitivities to design decisions remains. 

\begin{figure}
    \centering

    \includegraphics[width=\textwidth]{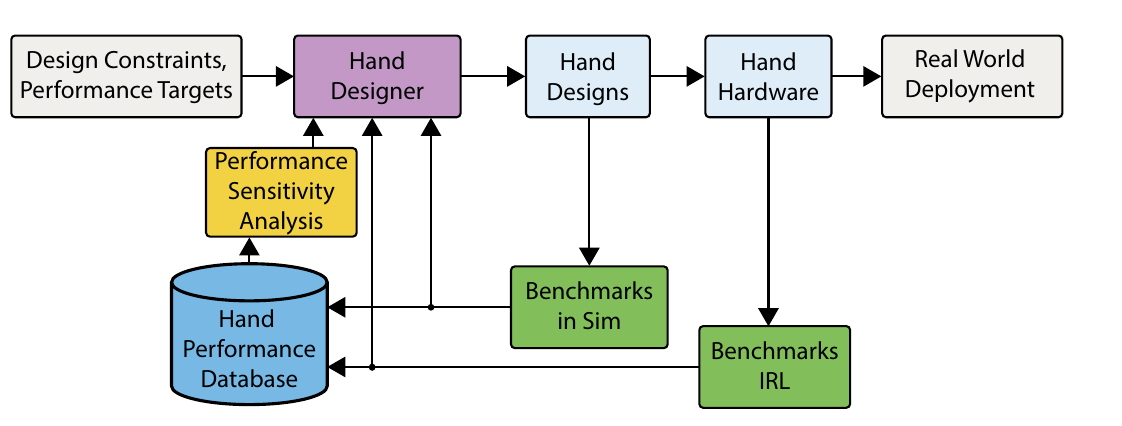}
    \caption{Benchmarks should track progress and provide actionable feedback to designers.}

    \label{fig:DesignFeedback}
\end{figure}

In the absence of a comprehensive design theory for multifingered hands, an empirical approach to establishing performance sensitivities, based on simulation and rapid prototyping, is summarized in Figure~\ref{fig:DesignFeedback}.
We envision design feedback loops that build a database of component-, hand-, and system-level benchmark performance and sensitivities using (1) simulated benchmarks of a large number of robot hand designs and (2) a much smaller number of experimental benchmarks of physical hand designs.

A major research challenge is developing simulations that can accurately account for contact-rich interactions~\cite{salvato_crossing_2021}, including hard contacts (e.g., the transition from the lid's six degrees of freedom to a single degree of freedom during threading on a jar) and soft contacts (e.g., during a sock-bundling task). It is also difficult to simulate the complex behavior of soft skins and tendon transmissions~\cite{baggetta_modeling_2024}. 
These issues are sometimes addressed in simulation using domain randomization, but there is a critical need for efficient simulators that can better model robot hands in real environments, including tendons, soft skin, tactile sensors, and multiple simultaneous hard and soft contacts.

Rapid prototyping enables fast build and test cycles~\cite{zhang2026modular}, but materials used for rapid prototyping often introduce excessive friction, compliance, hysteresis, creep, poor strength and durability, and other undesirable material properties that negatively impact mechanical transparency. New manufacturing techniques are needed to accelerate the physical design-build-test cycle of robot hands with high mechanical transparency. 

Despite these challenges, the field of dexterous robotics is moving faster than ever, due in part to unprecedented private investment. Appropriate benchmarking, as reviewed in this paper, will play a critical role in both measuring this progress and accelerating it.

%Disclosure
\section*{DISCLOSURE STATEMENT}
The authors are not aware of any affiliations, memberships, funding, or financial holdings that
might be perceived as affecting the objectivity of this review. 

% Acknowledgements
\section*{AUTHOR CONTRIBUTIONS}
AS led the overall manuscript preparation, including reference compilation, drafting, figure generation, and editing of the full manuscript, working with KML. Section~1 was drafted by KML. Section~2 was drafted by AR. 
Section~3 was drafted by SU with contributions and editing from JEC, LB, AB, and ST. Section~4 was drafted by AS and PR with contributions and editing from CM, LG, AAO, YY, AP, MGH, DG, RK, GCT, GKF, LB, AB, and ST. Section~5 was drafted by AS. 

% Acknowledgements
\section*{ACKNOWLEDGMENTS}
This material is based upon work supported by the National Science Foundation under Grant No.~2330040. We thank Cynthia Hipwell, Rob Ambrose, Brandon Krick, Toby Buckley, Raphael Cherney, and Shobhit Aggarwal for their valuable contributions during the preparation of this paper.

% References
\bibliographystyle{AR-Style/ar-style3}
\bibliography{
    References/Intro,
    References/Discussion,
    References/HighlevelBenchmarking,
    References/HandLevel,
    References/ComponentLevel
}

\end{document}